\documentclass{bmvc2k}
\usepackage{amsmath}
\usepackage{amssymb}
\usepackage{multirow}
\usepackage[flushleft]{threeparttable}
\usepackage{booktabs}
\usepackage{algorithm}
\usepackage{algpseudocode}
\usepackage{subfloat}
\usepackage{mathtools, nccmath}

\title{PCAS: Pruning Channels with Attention Statistics for Deep Network Compression}

\addauthor{Kohei Yamamoto}{yamamoto833@oki.com}{1}
\addauthor{Kurato Maeno}{maeno284@oki.com}{1}

\addinstitution{
 Oki Electric Industry Co., Ltd., Japan
}

\runninghead{Yamamoto, Maeno}{Pruning channels with attention statistics}

\def\eg{\emph{e.g}\bmvaOneDot}

\def\etal{\emph{et al}\bmvaOneDot}

\DeclareMathOperator*{\argmin}{arg\,min}
\newcommand{\bftab}{\fontseries{b}\selectfont}

\begin{document}

\maketitle
\begin{abstract}
Compression techniques for deep neural networks are important for implementing them on small embedded devices.
In particular, channel-pruning is a useful technique for realizing compact networks.
However, many conventional methods require manual setting of compression ratios in each layer.
It is difficult to analyze the relationships between all layers, especially for deeper models.
To address these issues, we propose a simple channel-pruning technique based on attention statistics that enables to evaluate the importance of channels.
We improved the method by means of a criterion for automatic channel selection, using a single compression ratio for the entire model in place of per-layer model analysis.
The proposed approach achieved superior performance over conventional methods with respect to accuracy and the computational costs for various models and datasets.
We provide analysis results for behavior of the proposed criterion on different datasets to demonstrate its favorable properties for channel pruning.
\end{abstract}
\vspace{-15pt}
\section{Introduction}
\vspace{-8pt}
{\it Convolutional neural networks} (CNNs) have brought about great advances in tasks such as object recognition, object detection, and semantic segmentation in several years.
However, the number of parameters required for CNNs that have generally good performance tends to be very large, which imposes memory requirements and computational cost that exceed the capabilities of mobile and compact devices.
To solve the problems, various techniques \cite{NIPS2014_5544,He2016DeepRL,Hinton2015DistillingTK,Hubara2016BinarizedNN,Szegedy2015GoingDW} have been proposed for making CNNs more efficient and increasing the speed of inference.
In these works, network pruning is an important approach for removing redundant parameters from the models.
\par Research into the pruning methods are roughly divided at two levels: the neuron level and the channel level.
At the neuron level, the number of parameters is reduced by severing connections between spatial neurons in the convolutional layer or the neurons in the fully connected layer.
At the channel level (channel pruning), the connections of all structural elements that respond to a particular channel are dropped for input and output channels in the convolutional layer; pruning is performed in sets of groups.
This method differs from the neuron-level pruning (\eg, in \cite{han2015deep}) in that it does not require any special implementation since the shape of the weight matrix is reduced.
However, since deletions are performed in sets of groups, the influence on the precision is significant and the problem setup is more difficult than with neuron-level methods.
\begin{figure}[t]
  \setlength{\belowcaptionskip}{-20pt}
  \setlength{\abovecaptionskip}{-10pt}
  \begin{center}
    \includegraphics[width=0.8\textwidth]{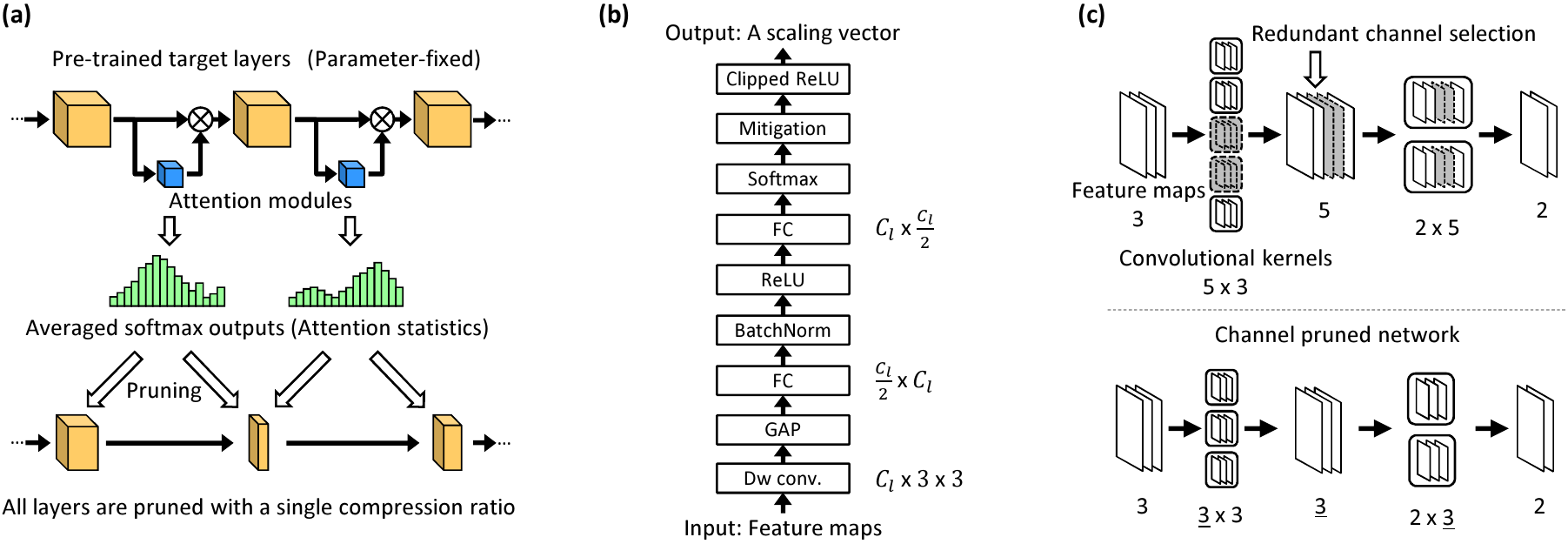}
    \caption{\small
    {\bf (a)} Overview of our pruning approach.
    {\bf (b)} The building blocks of a single attention module (see details in Section \ref{sec:strategy}).
    {\bf (c)} An example of channel pruning for convolutional layers.}
    \label{fig1}
  \end{center}
\end{figure}
The channel pruning methods have several difficulties that require designing the criteria for evaluating the importance of channels and set the compression ratio for each layer.
Especially, the latter is serious problem because the many existing methods~\cite{He2018SoftFP,he2017channel,iccv2017ThiNet,Li2016PruningFF,Yu_2018_CVPR} need the ratios as hyper-parameters for the pruning.
In general, the problem will be more difficult when using deeper models.
\par In this paper, we propose a channel pruning method for pre-trained models.
Figure \ref{fig1}a shows an overview of our approach.
In this method, the importance of channels is evaluated using neural networks (we call attention modules) connected immediately before all target layers in the pre-trained model.
Although these attention modules need to be trained, the modules are able to infer the importance of the channels.
Furthermore, it is optimized in all levels of layers since the attention module for a lower level is trained by considering the gradient of the pre-trained model and the gradient of the upper-level attention module.
\par The major contributions of this paper are summarized as follows:
\vspace{-5pt}
\begin{itemize}
  \setlength\itemsep{-3pt}
  \item
  We propose the attention statistics, a novel attention based criterion for channel pruning, to estimate redundant channels via optimizing the appended neural networks.
  \item
  We present a simple pruning technique that it requires only one compression ratio, which does not require the layer-by-layer compression ratio tuning that needs carefully controlling the trade-off between accuracy and the pruning performance.
  \item
  We evaluate our approach on various networks, VGG-10/16, ResNet-18/50/56, MobileNet and SegNet for image recognition/segmentation tasks.
  And the results show promising pruning performances on CIFAR-10/100, ImageNet and CamVid datasets.
\end{itemize}
\vspace{-15pt}
\section{Related Work}
\vspace{-20pt}
~~~~\par {\bf Non-pruning approaches.} Network quantization \cite{Hubara2016BinarizedNN, Rastegari2016XNORNetIC, Wan_2018_ECCV} is a technique for replacing typical 16/32-bit weights/activations with fewer-bit ones.
Hubara \etal~\cite{Hubara2016BinarizedNN} proposed a training scheme of binarized networks and Rastegari \etal~\cite{Rastegari2016XNORNetIC} improved it by introducing scaling factors to minimize quantization error.
Wan \etal~\cite{Wan_2018_ECCV} incorporated binary weight and ternary input to achieve better performance.
Knowledge distillation \cite{Hinton2015DistillingTK, ANC} is applied to train a small student model using a larger teacher one.
Belagiannis \etal~\cite{ANC} presented a two-player adversarial learning scheme to train the student model.
Factorization \cite{Chollet2017XceptionDL, Howard2017MobileNetsEC} is an approach whereby a standard convolution is factored into more efficient operations.
Chollet~\cite{Chollet2017XceptionDL} developed a depth-wise separable convolution and Howard \etal~\cite{Howard2017MobileNetsEC} used it to design more efficient models with a width and resolution multiplier.
Although such non-pruning approaches are based on different perspectives, they can be combined with the channel pruning approach to obtain even more compact models.
~~~~\par {\bf Channel pruning.} Li \etal~\cite{Li2016PruningFF} performed selection of redundant channels using $\ell_1$-norms of per-channel weights.
To decide the compression ratio for each layer, they analyzed the precision degradation depending on the number of channels that were deleted.
However, since compression ratios were decided by the user, they were not necessarily optimal.
As for using the norm of per-channel weights as a criterion, He \etal~\cite{He2018SoftFP} proposed the soft-pruning approach that the less $\ell_2$-norm kernels were zeronized for each epoch during training (allowing for updating from zeros in the next epoch).
Luo \etal~\cite{iccv2017ThiNet} found redundant channels by using the reconstruction error of each layer as the criterion, which compared the output before and after excluding certain channels and identifies channels with smaller error as more important.
Since channels were selected in each layer, the relationships between layers could not be considered and the compression ratios needed to be set manually.
Furthermore, a lot of time was needed for fine-tuning each layer.
Yu \etal~\cite{Yu_2018_CVPR} were focused on the reconstruction error of the last layer before classification and estimated the less important neurons in the backward propagation of the scores that were derived from the error.
Similarly, He \etal~\cite{he2017channel} solved the optimization problem of minimizing the reconstruction error in each layer by assigning a variable to each channel as a method that used $\ell_1$ regularization.
Huang \etal~\cite{Huang_2018_ECCV} introduced additional scaling factors to not only the output of channels but also the residual branches, and trained them to close 0 with the sparsity regularization for pruning, as in \cite{he2017channel}.
Both Huang \etal~\cite{Huang2018LearningTP} and He \etal~\cite{He_2018_ECCV} used the reinforcement learning for channel pruning.
They pruned unimportant channels selected by the agent networks that were trained to maximize the specialized reward functions for improving pruning performance.
Furthermore, their methods had the property of the automatic channel selection in the same manner as for our proposed method.
\par {\bf Attention.} The attention mechanism~\cite{NIPS2014_5542} that explicitly propagates positions to reference in spaces or in series are used for several applications.
Recent image recognition research have worked to increase accuracy by applying the mechanism.
Wang \etal~\cite{Wang2017ResidualAN} applied attention in spatial and channel directions for ResNet~\cite{He2016DeepRL}.
Hu \etal~\cite{hu2018senet} introduced attention in only channel directions to increase the performance of recognizing features by emphasizing channels according to the input.
The application of the mechanism to the model optimization or pruning has not yet been common.
\vspace{-10pt}
\section{Approach}
\vspace{-6pt}
In this section, we first provide a brief background on channel pruning.
Then we discuss the whole scheme of our approach and its details (including the pruning criterion, the training trick and the technique for selecting redundant channels).
\par {\bf Background.} In CNNs, the convolutional kernel (or filter) for the $l \in \{ 1, \ldots, L\}$-th layer is represented by a fourth-order tensor of the dimension $C_{l+1}\times C_l\times H_l\times W_l$, where $C_l$ is the number of channels, and $H_l$ and $W_l$ are the width and height of the kernel, respectively.
Note that $C_{l+1}$ belongs to the output side when $C_l$ belongs to the input side.
In general channel pruning schemes as shown in Fig. \ref{fig1}c, redundant channels are first removed by some kind of criterion.
By removing several channels from the feature maps in this way, the dimensions $C_{l+1}$ and $C_l$ of the corresponding convolutional kernels can also be removed.
After removing part of the channels, damage from pruning can be restored through fine-tuning of the model using the training data.
Although our approach also follows this scheme, we introduce a new strategy for the selection of redundant channels.
\vspace{-10pt}
\subsection{Pruning Strategy with Attention} \label{sec:strategy}
\vspace{-6pt}
Our goal is to estimate redundant channels precisely in pre-trained CNNs using the other neural networks (attention modules).
First, as visualized in the Fig. \ref{fig1}a, we connect the attention modules to immediately before the all of the pre-trained layer.
Next, we train the modules without updating any of parameters of the pre-trained layers under the same conditions (the same training data and the same loss function are used).
After training, we calculate our proposed pruning criterion (details in Sec. \ref{sec:criterion}) from the modules, and convert a given single compression ratio into the per-layer compression ratios using that criterion (details in Sec. \ref{sec:ratio}).
Then we determine the redundant channels in the pre-trained layer based on both the criterion and the compression ratios, and prune them.
Finally, we remove all the modules and then fine-tune the pruned network to restore pruning damage.
\par The role of the attention modules is to emphasize channels that contribute to reducing the loss function and to deemphasize others.
Because deemphasizing a channel produces the same effect as deemphasizing a group of convolutional weights that correspond to the channel, we assume that the deemphasized groups of weights have less influence on the original network and can be pruned with less accuracy degradation.
More specifically, the attention module generates a $C_l$-dimensional vector from the feature maps, and emphasis is achieved by a channel-wise multiplication of the vector by the feature maps again.
This operation is also known as the self-gating or the self-attention mechanism~\cite{hu2018senet,Wang2017ResidualAN}.
We expect that this mechanism can also be useful for evaluating the importance of channels.
\par Fixing the pre-trained weights causes the attention module to search for a solution that reduces the loss function under conditions of only being able to scale each of the input channels to nearly all layers of the pre-trained model.
Although this means that multiple attention modules are connected when there are three or more convolutional layers in the pre-trained model, the training of these is performed simultaneously.
Thus, the lower-layer attention module performs optimization based on the gradients both from the pre-trained model and from the upper-layer attention module.
In other words, the attention layers are optimized overall since the relationships with other layers are considered.
\par We now describe the architecture of the attention modules as shown in Fig. \ref{fig1}b.
First, depth-wise convolution is executed independently on the channels~\cite{Chollet2017XceptionDL} for extracting spatially common feature maps from the parameter-fixed layer outputs.
Next, global average pooling (GAP)~\cite{Lin2013NetworkIN}, Fully connected (FC) layer, batch normalization~\cite{Ioffe2015BatchNA}, and a ReLU function are applied to emphasize the difference between channels~\cite{hu2018senet}. Finally, the softmax function, mitigation function multiplication (details in Section \ref{sec:training}), and clipped ReLU~\cite{Hannun2014DeepSS} are applied.
\vspace{-20pt}
\subsection{Pruning Criterion} \label{sec:criterion}
\vspace{-6pt}
\par Naturally, since the behavior of attention varies according to the input data, it cannot be used as-is as a channel pruning criterion.
We therefore propose the attention statistic that is a quantity found by element-wise averaging of the softmax outputs of the attention modules over all training data, as a criterion for selecting important channels.
It is defined as follows:
\begin{equation}
  a_{l,c} = \frac{1}{|\mathcal{D}|}\sum_{i \in \mathcal{D}} s_{l,c,i}, \label{eq1}
\end{equation}
where $s_{l,c,i}$ is the output of the softmax function, $c \in \{1, \ldots, C_l\}$ is the index of the channel, and $\mathcal{D}$ is the set of training data.
Furthermore, we take $a_{l,c} \in \mathcal{A}_l$, as described below.
$\mathcal{A}_l$ is a set of channels in the attention statistics for the $l$-th layer, and $|\mathcal{A}_l|=C_l$.
In fact, the difference between channels needs to be emphasized.
For example, in~\cite{hu2018senet,Wang2017ResidualAN}, although the sigmoid independently to each channel is used to construct the scaling vector, it is difficult to obtain a clear difference in the statistical quantity when comparing between channels.
In the case of adjusting the inputs to the pre-trained model, this is because when the pre-trained model is assumed to be fully optimized, using nearly 1 for all values gives the best accuracy.
We therefore introduce the softmax function as a constraint that emphasizes the difference between channels.
Due to the property of the softmax output, emphasizing certain channels requires deemphasizing others.
This is a desirable property for emphasizing difference between channels.
\vspace{-12pt}
\subsection{Training the Attention Modules} \label{sec:training}
\vspace{-6pt}
Training the attention modules might perform poorly if the output of the softmax function is applied to the input feature maps as-is.
This is because the constraints on the pre-trained model with fixed parameters are too tight.
If we assume that all of the channels of the input feature maps that belong to the $l$-th layer have the same importance and the attention module is able to perfectly infer this, then the element values of output will all be $1/C_l$.
In other words, the feature maps become smaller depending on the magnitude of $C_l$ after multiplication.
Now, since the gradient is kept low each time a module spans across multiple layers, the gradient of the attention modules near the input layer disappears.
\par We therefore relax this constraint by multiplying the mitigation function to the output elements of the softmax function.
The mitigation function can be defined as
\begin{equation}
  f_l(\alpha) = \frac{C_l}{1+\alpha(C_l-1)}, \label{eq2}
\end{equation}
where $\alpha \in [0,1]$ is a hyper-parameter for strength of the constraint on the softmax.
If we assume that all the output elements of the attention module are $1/C_l$ when $\alpha=0$, then all of the channels of the feature maps are completely unaffected by the attention module (i.e., an identity projection), and the gradient can also propagate without decreasing.
However, in the case of $\alpha \to 1$, it approaches the output from the softmax alone.
The important effect is that the value of $\alpha$ gradually increases each time the attention module weights are updated to strengthen the constraints.
This effect allows the loss divergence to be avoided and solutions emphasizing the channel difference to be obtained.
Although performance deteriorates from the original pre-trained model as $\alpha$ is increased, since the aim of this process is for training the attention modules, this degradation is not a problem as long as the optimization is successful under the given constraints.
For stable learning, we provide the same range ($[0,1]$) as the softmax by applying the clipped ReLU~\cite{Hannun2014DeepSS}.
It can prevent the loss value from increasing fairly in the case that a scaling vector heavily enlarges specific channels.
\par This approach can also be applied to the FC layers as a structured pruning method.
A pruning evaluation for the FC layers is also included in Section~\ref{sec:experiments}.
\vspace{-12pt}
\subsection{A Single Compression Ratio for All Layers}\label{sec:ratio}
\vspace{-6pt}
Our method only needs a single compression ratio $r \in [0, 1]$ that can be converted into the ratio for each layer $l$.
Note that the ratio $r$ is proportional to the total number of channels contained in all layers.
First, we define the global threshold $t \in [0, T]$ where $T \in \mathbb{R}_{>0}$ is an arbitrary number, and the local threshold $t/C_l$.
There is a clear relationship that the global threshold can also be converted into the local one.
Here, the total number of channels whose attention statistics are under the local threshold can be represented as follows:
\begin{equation}
  U_l(t) = |\{a \colon a < t/C_l \cap a \in \mathcal{A}_l \}|. \label{eq3}
\end{equation}
Next, we define the local compression ratio $U_l(t)/C_l$ for each layer $l$ and the global compression ratio $g(t)=(\sum_{l=1}^{L} U_l(t)) / (\sum_{l=1}^{L} C_l)$.
The optimal global threshold $t^*$ is found by solving the problem formulated as follows:
\useshortskip
\begin{equation}
  t^* = \argmin_{t} \left|~ g(t) - r ~\right|. \label{eq4}
\vspace{-8pt}
\end{equation}
Since it has only one parameter $t$ and the convexity property, search methods (\eg, the grid search) can solve this problem easily.
Finally, the $l$-th layer's channels whose criteria are under $t^*/C_l$ are selected as the pruning targets.
We expect the softmax distribution to flatten as the number of important channels increases.
This method aims to prevent important channels with flat distributions from being pruned and redundant channels with non-flat distributions from remaining.
In general, when the criterion is used for evaluating the importance of channels, it is difficult to directly compare channels that belong to different layers.
Although attention statistics are also not strictly comparable with those channels, we propose a roughly fair comparison technique using the properties of per-layer normalization by the softmax function.
In many cases, important channels are kept even if the $r$ value is higher.
If all channels could be pruned in a layer, just keeping the most important channel is a simple workaround.
\vspace{-8pt}
\section{Experiments}
\vspace{-8pt}
In this section, we first evaluated effectiveness in comparison with conventional state-of-the-art pruning methods against the various models.
Next, we report the ablation study results.
Finally, we analyze the behavior of the attention statistics.
\par We evaluated the proposed method against the CIFAR-10/100~\cite{Krizhevsky2009LearningML} and the ImageNet (ILSVRC-2012)~\cite{Deng2009ImageNetAL} datasets for the object recognition task, and the CamVid road scenes~\cite{Brostow2009SemanticOC} dataset for the semantic segmentation task.
In all experiments, we used the SGD optimization algorithm with a momentum of $0.9$ to train the attention modules and to fine-tune the pruned networks, and did not prune the first convolutional layer, where the influence was wide and significant.
As when pruning the ResNet-type architecture in all datasets, a sampling technique~\cite{he2017channel} for discarding an arbitrary number of input channels at the start of the residual branch was introduced to expand the range of target channels.
We implemented our proposed method on Chainer~\cite{chainer_learningsys2015}.
Note that all evaluated models in these experiments had 32-bit floating point weights.
The experimental settings for each dataset are described below.
\par{\bf CIFAR-10/100.}~~We evaluated VGG-10 and ResNet-18/56 on CIFAR-10, and evaluated VGG-10 and ResNet-50 on CIFAR-100.
The attention modules were trained for 50 epochs with a learning rate of $10^{-2}$ and the value $\alpha$ was linearly increased from $0$ to $6\times10^{-2}$, and then the rate was changed to $10^{-3}$ for another 50 epochs of training to stabilize the solution with the target value of $\alpha$ using a batch size of 128.
We adopted some data augmentation techniques: horizontal flip, image expansion~\cite{Liu2016SSDSS}, and random crop (where images are cropped to $28~\times~28$ pixels).
\par{\bf ImageNet (ILSVRC-2012).}~~We evaluated our pruning method on VGG-16~\cite{Simonyan2014VeryDC}, ResNet-18/50~\cite{He2016DeepRL} and MobileNet~\cite{Howard2017MobileNetsEC}.
During training of attention modules for all models, the value $\alpha$ was linearly increased from $0$ to $2\times10^{-3}$ for 5 epochs, then training continued for 5 more epochs for stabilization.
The learning rate was fixed at $10^{-3}$ while training the attention modules.
In the fine-tuning step, VGG-16 and MobileNet were trained for 45 epochs, dropping the learning rate from the initial $10^{-3}$ value by $10^{-1}$ every 15 epochs.
In contrast, ResNet-18/50 were trained for 35 epochs and the learning rate was changed from $5\times10^{-3}$ to $5\times10^{-5}$ in the same manner as for VGG-16.
We used batch sizes of 512 for VGG-16, ResNet-18 and MobileNet, and of 1024 for ResNet-50.
We adopted two standard data augmentation techniques: $224\times224$ random cropping and horizontal flip.
\par{\bf CamVid.}~~We evaluated with the SegNet~\cite{Badrinarayanan2017SegNetAD} architecture.
We trained the attention modules while increasing the value $\alpha$ from $0$ to $2\times10^{-3}$ and with a learning rate of $10^{-2}$ for 50 epochs, then trained with a fixed $\alpha$ and a learning rate of $10^{-3}$ for 50 more epochs using a batch size of 8.
We adopted a data augmentation technique: horizontal flip.
\vspace{-12pt}
\subsection{Comparison with Conventional Methods} \label{sec:experiments}
\vspace{-8pt}
For comparisons, we measured the number of floating point operations (FLOPs\footnote{In this study, FLOPs indicate only the number of operations in the convolutional or FC layer. Furthermore, FLOPs were calculated without considering the fused multiply-add (FMA) instruction for comparison, except for ResNet-56 on CIFAR-10.}), and the total number of parameters.
We generated the pruned models with some compression ratios, and then selected the model that had the nearest accuracy to the compared method and noted the best pruning performance in the results.
\par {\bf Recognition model pruning.} Table~\ref{tb1} compares the proposed method with conventional methods in the object recognition task that is commonly used for pruning performance evaluations.
In Table~\ref{tb1}, M/B means $10^6$/$10^9$, and arrows indicate absolute accuracy reductions or reduced ratios for the number of parameters and FLOPs.
Our method is denoted as ``PCAS'', and the compression ratio usage is appended (\eg, ``-10'' indicates that $10\%$ of the compression ratio $r$ is used).
For ResNet-56 on CIFAR-10, PCAS outperformed the conventional methods that were based on the norm of weights~\cite{He2018SoftFP,Li2016PruningFF}, the reconstruction error~\cite{Yu_2018_CVPR} and the reinforcement learning~\cite{He_2018_ECCV}.
In the CIFAR-100 experiment, we used a ResNet-50 model whose reduction layers were replaced with zero-padding~\cite{Han2017DeepPR} to reduce the number of parameters.
Although ResNet-50 has more parameters than ResNet-56, PCAS reduced redundant channels to a greater extent compared with the conventional methods.
Regarding VGG-16, PCAS tended to reduce the number of the parameters more than other methods did, while also reducing the number of FLOPs at same levels of accuracy.
As for ResNet-50, PCAS outperformed the other methods that included the depth pruning approach~\cite{Huang_2018_ECCV} and achieved the network with the fewest parameters and FLOPs.
\begin{table*}[t]
\setlength{\tabcolsep}{4.25pt}
\centering
\caption{\small Comparison among several different pruning methods for the object recognition.}
\label{tb1}
\fontsize{6.5}{6}\selectfont
\begin{tabular}{@{}lllrlrlrlrl@{}}
\toprule
Dataset & Model & Method & \multicolumn{2}{l}{Top-1 Acc. $\%$} & \multicolumn{2}{l}{Top-5 Acc. $\%$} & \multicolumn{2}{l}{\#Params.} & \multicolumn{2}{l}{\#FLOPs} \\
\midrule
\multirow{5.5}{*}{CIFAR-10} & \multirow{5.5}{*}{ResNet-56}
& Pruned-B \cite{Li2016PruningFF} & $93.06$ & $\uparrow 0.02$ &-~~~~~~~&& $0.73$M & $\downarrow 13.7\%$ & $91$M & $\downarrow 27.6\%$ \\
&& NISP-56 \cite{Yu_2018_CVPR} &-~~~~~~~& $\downarrow 0.03$ &-~~~~~~~&& $0.49$M & $\downarrow 42.6\%$ & $71$M & $\downarrow 43.6\%$ \\
&& AMC \cite{He_2018_ECCV} & $91.90$ & $\downarrow 0.90$ &-~~~~~~~&&-~~~~~~~&& $63$M & $\downarrow 50.0\%$ \\
&& SFP (40$\%$) \cite{He2018SoftFP} & $93.35$ & $\downarrow 0.24$ &-~~~~~~~&&-~~~~~~~&& $59$M & $\downarrow 52.6\%$ \\
&& PCAS-35 & $93.58$ & $\uparrow$~\bftab{0.54} &-~~~~~~~&& $0.39$M & $\downarrow$~\bftab{53.7}$\%$ & $56$M & $\downarrow$~\bftab{54.8}$\%$ \\
\midrule
\multirow{2}{*}{CIFAR-100} & \multirow{2}{*}{ResNet-50}
& \cite{Li2016PruningFF} (our impl.) & $73.60$ & $\downarrow 0.86$ &-~~~~~~~&& $7.83$M & $\downarrow 54.2\%$ & $616$M & $\downarrow 56.3\%$ \\
&& PCAS-60 & $73.84$ & $\downarrow$~\bftab{0.62} &-~~~~~~~&& $4.02$M & $\downarrow$~\bftab{76.5}$\%$ & $475$M & $\downarrow$~\bftab{66.3}$\%$ \\
\midrule
\multirow{14}{*}{ImageNet} & \multirow{8}{*}{VGG-16}
& ThiNet-Conv \cite{iccv2017ThiNet} & $69.80$ & $\uparrow$~\bftab{1.46} & $89.53$ & $\uparrow$~\bftab{1.09} & $131.44$M & $\downarrow 5.0\%$ & $9.58$B & $\downarrow 69.0\%$ \\
&& PCAS-45 & $69.41$ & $\uparrow 1.00$ & $89.22$ & $\uparrow 0.85$ & $128.95$M & $\downarrow$~\bftab{6.8}$\%$ & $8.59$B & $\downarrow$~\bftab{72.2}$\%$ \\
\cmidrule{3-11}
&& SSS \cite{Huang_2018_ECCV} & $68.53$ & $\downarrow 3.93$ & $88.20$ & $\downarrow 2.64$ & $130.50$M & $\downarrow 5.6\%$ & $7.67$B & $\downarrow 75.2\%$ \\
&& PCAS-50 & $68.83$ & $\uparrow$~\bftab{0.42} & $88.82$ & $\uparrow$~\bftab{0.45} & $128.05$M & $\downarrow$~\bftab{7.4}$\%$ & $7.49$B & $\downarrow$~\bftab{75.8}$\%$ \\
\cmidrule{3-11}
&& CP (5$\times$) \cite{he2017channel} & $67.80$ &-~~~~~~~& $88.10$ & $\downarrow 1.80$ & $130.88$M & $\downarrow 5.4\%$ & $7.03$B & $\downarrow 77.2\%$ \\
&& PCAS-55 & $68.18$ & $\downarrow 0.23$ & $88.39$ & $\uparrow$~\bftab{0.02} & $127.23$M & $\downarrow$~\bftab{8.0}$\%$ & $6.45$B & $\downarrow$~\bftab{79.2}$\%$ \\
\cmidrule{2-11}
 & \multirow{5.35}{*}{ResNet-50}
& CP (2$\times$) \cite{he2017channel} & $72.30$ & $\downarrow 3.00$ & $90.80$ & $\downarrow 1.40$ & $17.46$M & $\downarrow 31.5\%$ & $5.20$B & $\downarrow 32.8\%$ \\
&& ThiNet-70 \cite{iccv2017ThiNet} & $72.04$ & $\downarrow 0.84$ & $90.67$ & $\downarrow 0.47$ & $16.94$M & $\downarrow 33.7\%$ & $4.88$B & $\downarrow 36.8\%$ \\
&& SSS-ResNet-26 \cite{Huang_2018_ECCV} & $71.82$ & $\downarrow 4.30$ & $90.79$ & $\downarrow 2.07$ & $15.60$M & $\downarrow 38.8\%$ &-~~~~~~~& $\downarrow 43.0\%$ \\
&& NISP-50-B \cite{Yu_2018_CVPR} &-~~~~~~~& $\downarrow 0.89$ &-~~~~~~~&& $14.36$M & $\downarrow 43.8\%$ & $4.32$B & $\downarrow 44.0\%$ \\
&& PCAS-50 & $72.68$ & $\downarrow$~\bftab{0.04} & $91.09$ & $\uparrow$~\bftab{0.03} & $12.47$M & $\downarrow$~\bftab{51.2}$\%$ & $3.34$B & $\downarrow$~\bftab{56.7}$\%$ \\
\bottomrule
\end{tabular}
\vspace{-15pt}
\end{table*}
\par {\bf Segmentation model pruning.} Table~\ref{tb2} ({\it left}) shows pruning performance for the semantic segmentation task using SegNet on CamVid.
PCAS showed competitive results.
We confirmed that PCAS was able to exceed a $10\%$ reduction in numbers of parameters compared with conventional methods that had the same level of accuracy and variations in FLOPs.
Although PCAS reduced more than $60\%$ of parameters with a relatively small compression ratio $30\%$, this indicated that most of the redundant channels had almost the same importance.
This result suggests the effectiveness of PCAS for architectures with many channels in the middle of layers and for this task.
\begin{table}[h]
  \centering
  \caption{\small Pruning SegNet on CamVid ({\it left}). Pruning the FC layers of VGG-16 on ImageNet ({\it right}).}
  \setlength{\tabcolsep}{3pt}
  \label{tb2}
\fontsize{6.5}{6}\selectfont
\begin{tabular}{lllll}
\toprule
Method     & \multicolumn{2}{l}{Global Acc. $\%$} & \#Params. & \#FLOPs \\
\midrule
\cite{Li2016PruningFF} (\cite{Huang2018LearningTP}'s impl.) & 83.50 & $\downarrow 3.00$ & $\downarrow 56.9\%$ & $\downarrow$~\bftab{63.9}$\%$ \\
LTP \cite{Huang2018LearningTP} & 88.60 & $\uparrow$~\bftab{2.10}& $\downarrow 56.9\%$ & $\downarrow$~\bftab{63.9}$\%$ \\
PCAS-30 & 88.57 & $\uparrow 0.82$ & $\downarrow$~\bftab{67.8}$\%$  & $\downarrow 63.8\%$ \\
\bottomrule
\end{tabular}
~~~
\begin{tabular}{lllll}
\toprule
Method     & \multicolumn{2}{l}{Top-1 Acc. $\%$} & \#Params. & \#FLOPs \\
\midrule
ThiNet-GAP \cite{iccv2017ThiNet} & 67.34 & $\downarrow 1.0$ & $\downarrow$~\bftab{94.0}$\%$ & $\downarrow 69.8\%$ \\
PCAS-L-55 & 67.91 & $\downarrow$~\bftab{0.5} & $\downarrow 83.0\%$  & $\downarrow$~\bftab{72.9}$\%$ \\
\bottomrule
\end{tabular}
\vspace{-15pt}
\end{table}
\par {\bf FC layer pruning.} PCAS could be applied to structured pruning for FC layers by removing the depth-wise convolutional layer and the GAP operation from the attention modules.
For ImageNet, we also evaluated this approach using VGG-16, which has numerous parameters in FC layers.
Due to conceptual differences between the channels, we took a two-step pruning approach. Namely, we pruned an already pruned network using the same training methods.
We chose the ``PCAS-45'' model in Table~\ref{tb1} as the target pruned network.
As Table~\ref{tb2} ({\it right}) shows, PCAS (with appended suffix ``-L'') was competitive with the GAP approach using conventional methods~\cite{iccv2017ThiNet}.
Although the GAP approach reduced the larger number of parameters by removing FC layers, it is not capable to control the trade-off between accuracy and computational costs, unlike the pruning approach.
\par {\bf Comparison with non-pruning approaches.} Table~\ref{tb3} shows the recognition results in comparison with the non-pruning compression approaches, including quantization, distillation and factorization techniques.
Among the ResNet-18 models, PCAS with a compression ratio of $25\%$ (PCAS-25) was superior to the quantization \cite{Rastegari2016XNORNetIC, Wan_2018_ECCV} and distillation \cite{ANC} approaches in terms of both the number of parameters and the accuracies.
In contrast, quantization methods have potential for faster inference speed and lower memory consumption due to the utilization of more efficient bitwise operations through a dedicated implementation.
MobileNet \cite{Howard2017MobileNetsEC} with settings of 0.75 channel-width multipliers was more compact than the PCAS-25 model.
Although MobileNet consists of depth-wise and 1 x 1 convolutional layers, PCAS can be also applied to such compact architectures by selecting only 1 x 1 convolutional layers as the channel pruning targets because of the channel independency of depth-wise convolutional layers.
For MobileNet, we experimented with $r=5\%, 10\%, 15\%$ (denoted as ``+ PCAS'') and it reduced about $15\%$ of parameters and $30\%$ of FLOPs with an accuracy degradation of $1.08\%$.
We found that accuracy for MobileNet tends to be more sensitive to the compression ratio than do larger models (\eg, VGG-16).
\vspace{-8pt}
\begin{table}[h]
  \centering
  \caption{\small Performance comparison with non-pruning approaches on ImageNet.}
  \setlength{\tabcolsep}{3pt}
  \label{tb3}
\fontsize{6.5}{6}\selectfont
\begin{tabular}{llcrrr}
\toprule
Method & Model & Top-1/5 Acc. $\%$ & \#Params. & \#FLOPs \\
\midrule
Original & ResNet-18 & 68.98 / 88.69 & 11.68M & 3.63B \\
XNOR-Net \cite{Rastegari2016XNORNetIC} & ResNet-18 & 51.20 / 73.20 & 11.68M & -~~~~ \\
TBN \cite{Wan_2018_ECCV} & ResNet-18 & 55.60 / 79.00 & 11.68M & -~~~~ \\
ANC \cite{ANC} & ResNet-18 & 67.11 / 88.28 & 13.95M & -~~~~ \\
PCAS-25 & ResNet-18 & 68.04 / 88.01 & 8.58M & 2.59B \\
\midrule
MobileNet \cite{Howard2017MobileNetsEC} (our impl.) & 0.75 MobileNet & 68.19 / 88.39 & 2.59M & 650M \\
MobileNet + PCAS-5 & 0.75 MobileNet & 68.11 / 88.35 & 2.45M & 571M \\
MobileNet + PCAS-10 & 0.75 MobileNet & 67.51 / 87.88 & 2.32M & 510M \\
MobileNet + PCAS-15 & 0.75 MobileNet & 67.11 / 87.58 & 2.19M & 458M \\
\bottomrule
\end{tabular}
\vspace{-15pt}
\end{table}
\begin{figure}[t]
  \setlength{\belowcaptionskip}{-15pt}
  \setlength{\abovecaptionskip}{-12pt}
  \centering
  \fontsize{5.8}{6}\selectfont
  \setlength\tabcolsep{0pt}
  \begin{tabular}{cc@{\hspace{-2pt}}c@{\hspace{-5pt}}c}
  \includegraphics[width=0.23\linewidth]{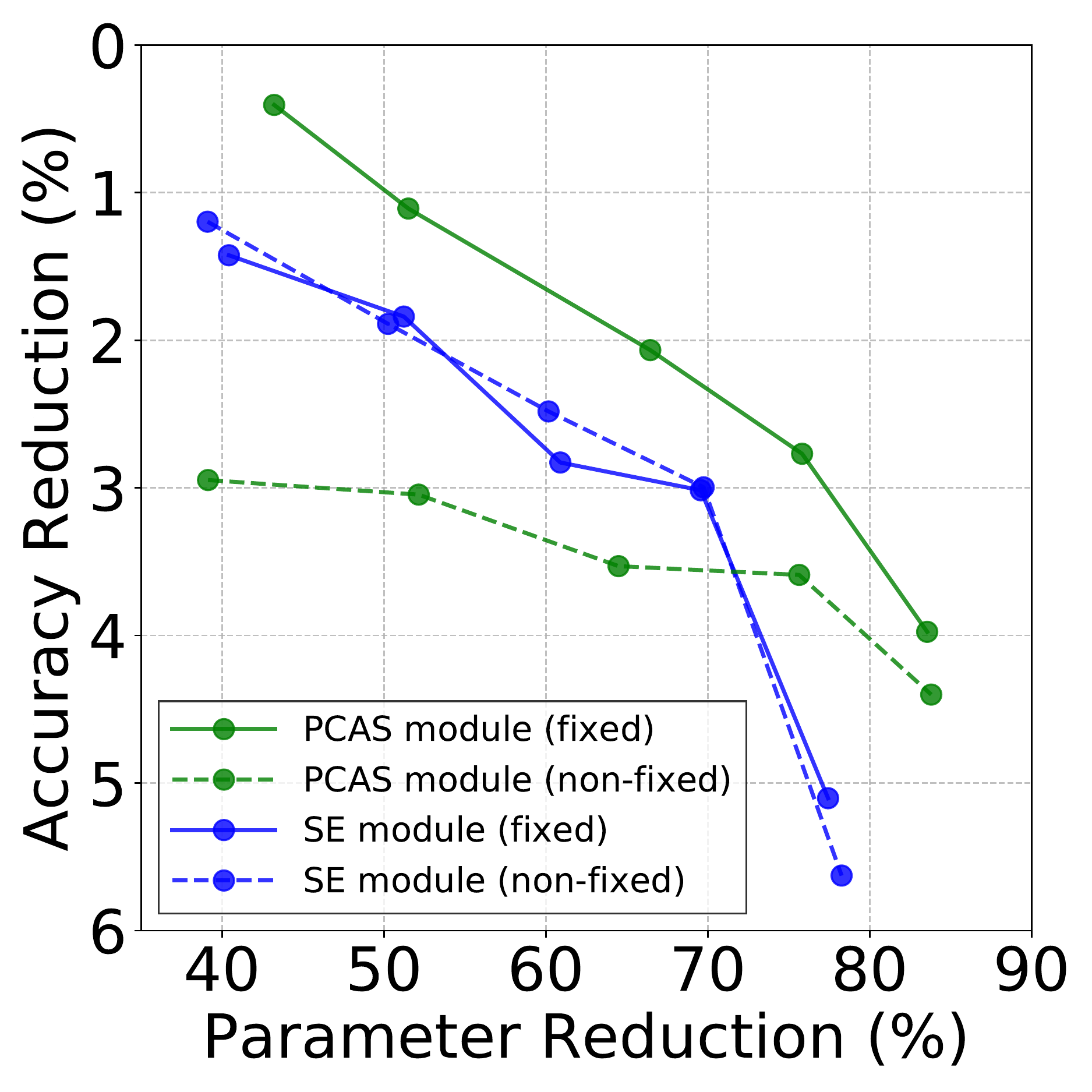}&
  \includegraphics[width=0.23\linewidth]{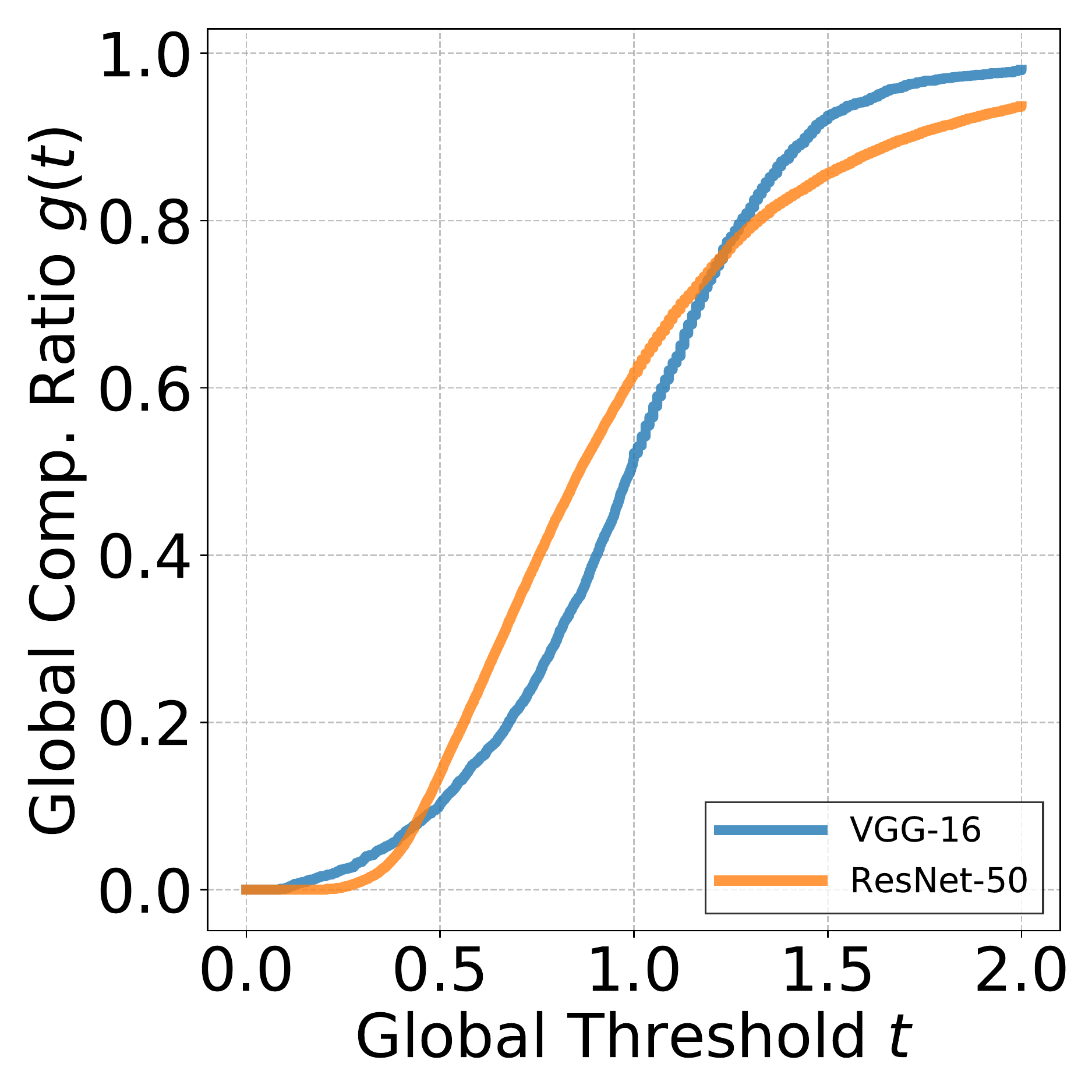}&
  \includegraphics[width=0.30\linewidth]{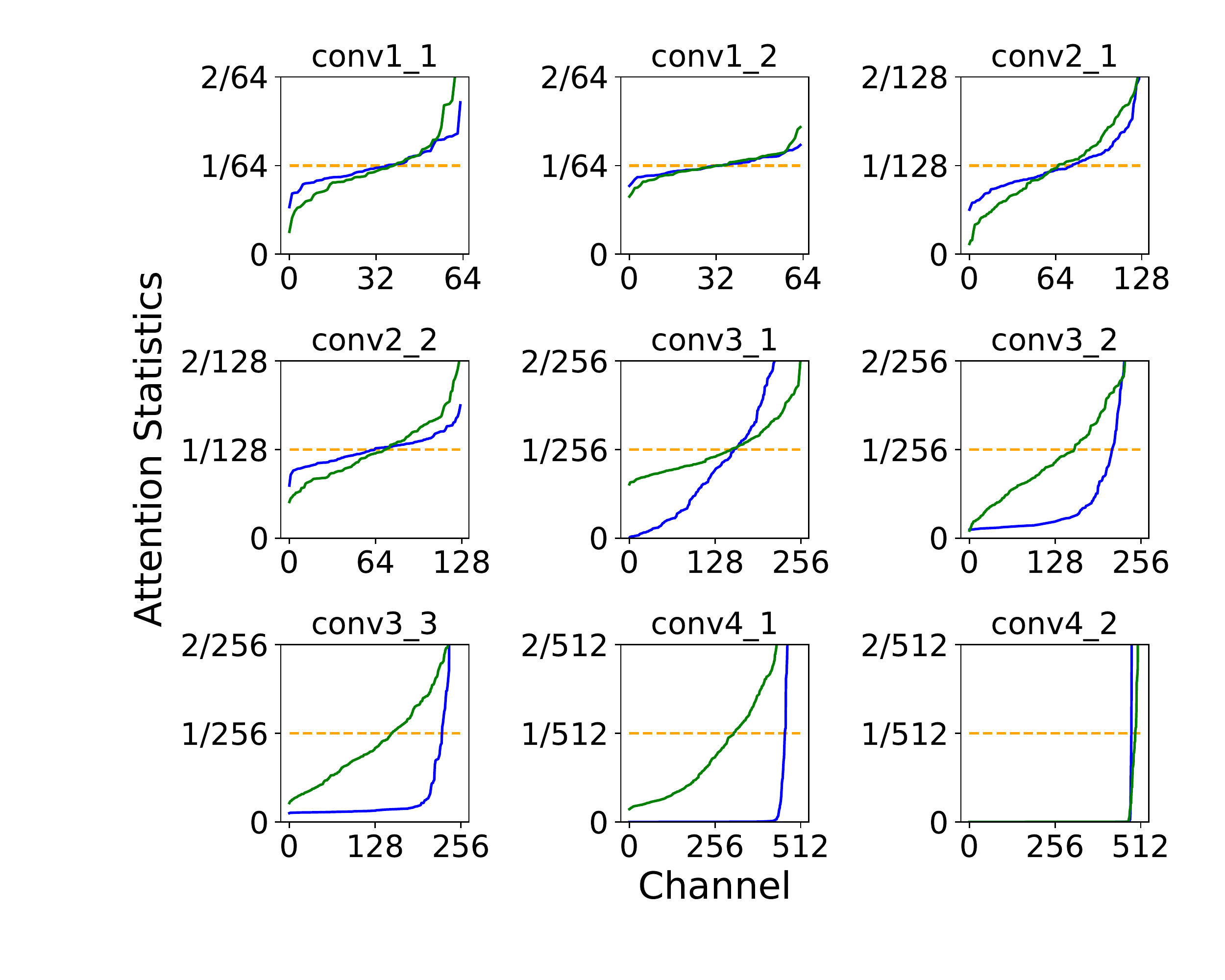}&
  \includegraphics[width=0.27\linewidth]{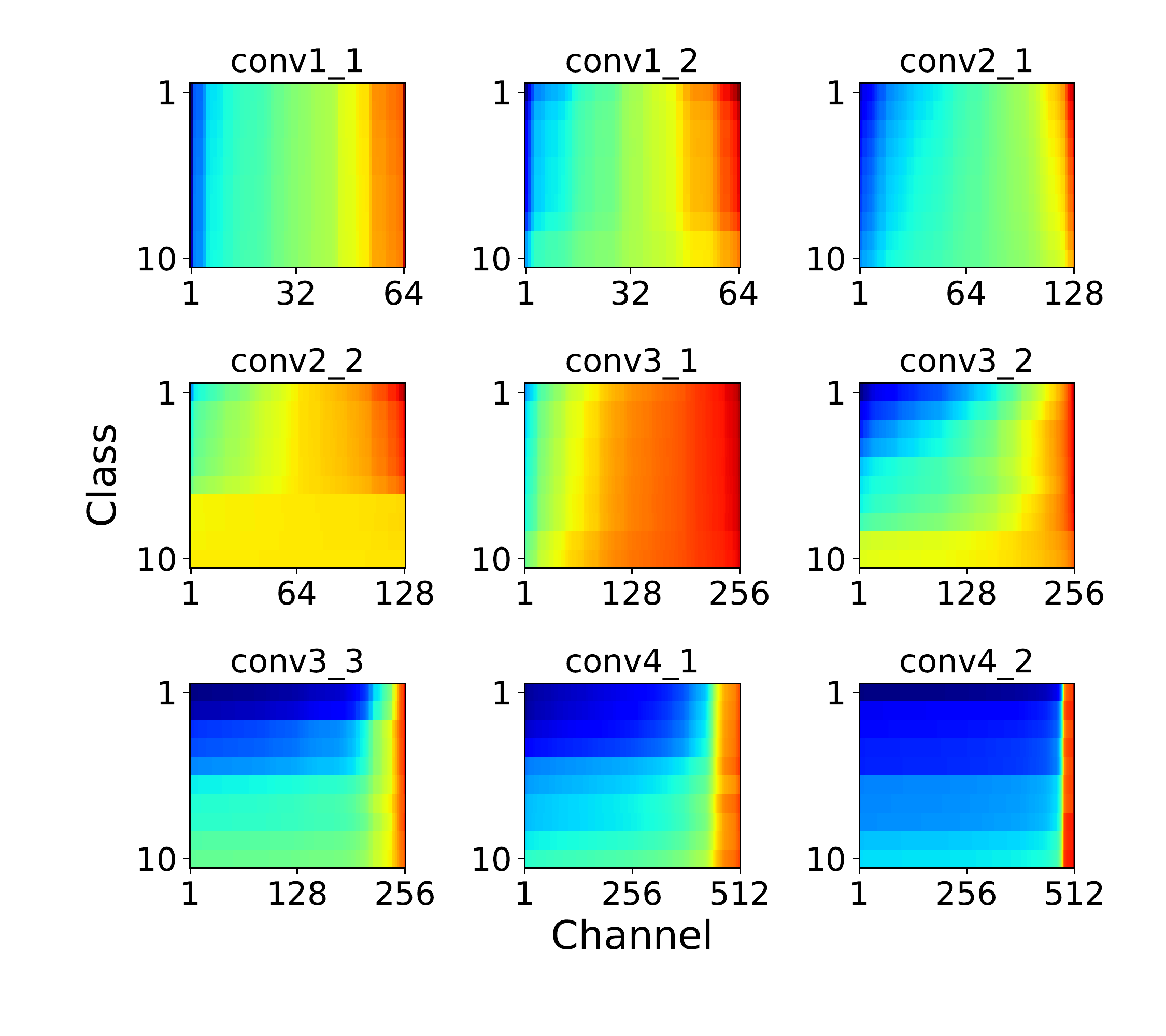}\\
  (a) ResNet-18 on CIFAR-10 & (b) VGG-16/ResNet-50 on ImageNet & (c) VGG-10 on CIFAR-10/100 & (d) VGG-10 on CIFAR-10
  \end{tabular}
  \caption{\small
  {\bf (a)} Performance comparison with the training schemes and the modules on ResNet-18.
  {\bf (b)} The relationship between $t$ and $g(t)$ in Eq. \ref{eq4} for VGG-16 and ResNet-50 on ImageNet.
  {\bf (c)} Attention statistics on CIFAR-10 (blue) and CIFAR-100 (green). Channels in each layer are sorted by their values.
  {\bf (d)} Attention statistics for each class.
  Values are converted to the log-scale for visibility and then colorized on each layer independently (red value is higher than blue).
  Channels and classes are sorted by the mean values over class axis and the minimum values over channel axis.
  }
  \label{fig2}
\end{figure}
\vspace{-6pt}
\subsection{Ablation Study}
\vspace{-6pt}
~~~~{\bf Parameter-fixed training.} We investigated the effectiveness of parameter-fixed training for attention modules, and differences in module architectures with conventional self-attention modules proposed in SE-Net~\cite{hu2018senet}.
The proposed method considers the importance of fixing parameters for the whole network without attention modules while training those for emphasizing the important differences between channels.
To confirm this relationship, we conducted the experiment using ResNet-18, whose reduction layers were replaced with zero-padding and CIFAR-10.
Figure~\ref{fig2}a shows the performance results after fine-tuning with regard to accuracy and parameter losses in four cases.
We denote PCAS as the use case of the proposed attention modules.
Clearly, the pattern in the non-fixed case was inferior to that in the fixed case.
Furthermore, we confirmed that actual attention statistics such as those shown in Fig.~\ref{fig2}a in the non-fixed case had small discrepancies that were not sufficiently emphasized in the fixed case.
From the above, we believe that this is because the weights in the original network are strongly dependent on the softmax constraint.
\par {\bf Attention module architecture.} We evaluated differences in the attention module architectures.
In contrast to the proposed modules using softmax, the SE modules in SE-Net independently apply the sigmoid function channel as the last blocks.
Unlike our method, we trained the appended SE modules in a straightforward way as long as the SE-module outputs had nothing to do with between-channel constraints as in softmax (i.e., the training rarely broke down).
The original network was then pruned using attention statistics, which were obtained from the sigmoid outputs.
From the results in Fig.~\ref{fig2}a, we confirmed that the SE modules were inferior to the parameter-fixed case of the proposed method.
We also found no large differences between fixed and non-fixed trainings using the SE module.
As the result, we consider that the procedure for mitigating the softmax constraints leads to effective extraction of channel importance.
\vspace{-5pt}
\begin{table}[h]
  \centering
  \caption{\small Pruning performance with different compression ratios on ImageNet.}
  \setlength{\tabcolsep}{3pt}
  \label{tb4}
\fontsize{6}{6}\selectfont
\begin{tabular}{crcrrcc}
\toprule
Model & $r$~~   & Top-1/5 Acc. $\%$ & \#Params. & \#FLOPs & Time (ms)\\
\midrule
\multirow{6}{*}{VGG-16}
& 0$\%$ & 68.41 / 88.37 & 138.34M & 30.94B & 71.41 \\
& 40$\%$ & 69.91 / 89.65 & 129.89M & 9.95B & 27.62 \\
& 45$\%$ & 69.41 / 89.22 & 128.95M & 8.59B & 22.13 \\
& 50$\%$ & 68.83 / 88.82 & 128.05M & 7.49B & 19.03 \\
& 55$\%$ & 68.18 / 88.39 & 127.23M & 6.45B & 16.94 \\
& 60$\%$ & 67.35 / 87.88 & 126.51M & 5.36B & 14.99 \\
\bottomrule
\end{tabular}
~~~~~~
\begin{tabular}{crcrrcc}
\toprule
Model & $r$~~   & Top-1/5 Acc. $\%$ & \#Params. & \#FLOPs & Time (ms)\\
\midrule
\multirow{6}{*}{ResNet-50}
& 0$\%$ & 72.72 / 91.06 & 25.56M & 7.72B & 55.46 \\
& 45$\%$ & 72.94 / 91.47 & 14.22M & 3.85B & 44.22 \\
& 50$\%$ & 72.68 / 91.09 & 12.47M & 3.34B & 42.91 \\
& 55$\%$ & 71.96 / 90.95 & 10.76M & 2.82B & 41.53 \\
& 60$\%$ & 71.23 / 90.44 & 9.08M & 2.37B & 39.91 \\
& 65$\%$ & 70.25 / 89.96 & 7.58M & 1.94B & 38.26 \\
\bottomrule
\end{tabular}
\vspace{-5pt}
\end{table}
\par {\bf Pruning with different compression ratios.} Figure \ref{fig2}b shows the relationship between $t$ and $g(t)$ in Eq. \ref{eq4} for both VGG-16 and ResNet-50 on ImageNet.
As mentioned in Section \ref{sec:training}, if all channels have the same importance for each layer, the relationship becomes the 0-1 step function switching at $t=1.0$.
However, since the curves slope upward, we confirmed that emphasizing the difference between channels was successfully achieved.
We therefore experimented with different compression ratios $r$ by 5\%, and the results are summarized in Table \ref{tb4}.
Note that the pruned models were generated from the same attention statistics, namely that the same curve in Fig. \ref{fig2}b was used for each model, and then they were fine-tuned under the same conditions.
``Time'' indicates the computational time of only a forward propagation, measured on a single GPU (NVIDIA Quadro GV100) with a batch size of 32.
Although the performance goes worse with increasing the $r$ value gradually, the pruned models gave better results than the original models till the $r$ value of $0.50$ and $0.45$ for VGG-16 and ResNet-50, respectively.
And it also shows our method reduced $60\%$ of channels with accuracy degradation of up to $1.5\%$ for both models.
\vspace{-8pt}
\subsection{Analysis of Attention Statistics}
\vspace{-6pt}
For analysis, we used VGG-10, which is composed of only ten convolutional layers of VGG-16 by replacing FC layers with GAP.
Figure~\ref{fig2}c shows nine attention statistics corresponding to the channels for dimensions $C_1=64$ to $C_9=512$ in each convolutional layer in VGG-10 on CIFAR-10/100.
These attention statistics directly show the importance of each layer.
For example, since {\it conv1\_2} has a shape that is closer to flat than other distributions, the importance of all of the channels is relatively high, and the effect on the accuracy is large.
Furthermore, the distribution in {\it conv4\_2} is heavily biased toward some channels, which indicates that there is a large number of redundant channels.
The structure of VGG-10 is set to have more channels nearer the output-layer side, and these are clearly redundant.
\par By comparing the results of CIFAR-10 and CIFAR-100, we confirmed that the redundancy contained in the trained VGG-10 differs depending on the complexity of the problem.
For example, from {\it conv4\_1}, it is clear that the CIFAR-10 distribution is more biased than CIFAR-100, and the importance of some channels is high.
In contrast, the CIFAR-100 distribution is only weakly biased, with other channels also contributing to the accuracy.
Since our method decides the threshold value for pruning based on the ratio to the number of channels, it does not emphasize the channels in {\it conv4\_1} for pruning in CIFAR-100 as much as in CIFAR-10.
As a result, our method prunes independently of the complexity of the problem.
\par Figure~\ref{fig2}d shows the attention statistics for each class.
This visualization shows that the response to particular classes is weak.
Furthermore, many of these are observed in output-side layers.
Since the average is used in our method, channels that have weak response over all classes are pruned preferentially over channels that have specifically weak response to particular classes.
\vspace{-8pt}
\section{Conclusion}
\vspace{-8pt}
In this paper, we proposed a novel pruning method based on attention.
Our goal was to more effectively extract the importance of channels from the original networks for pruning.
The method trains attention modules inserted immediately before the target pre-trained convolutional or FC layers to learn the importance of the channels.
After training, we can obtain pruning criteria from module inferences by taking statistics using the training data.
Attention modules are one-shot trained, not in a layer-by-layer manner.
We apply comparability to automate setting of the channel compression ratio for each layer in the entire model, whereas conventional methods set the ratio for each layer.
We conducted various experiments using VGG-10/16, ResNet-18/50/56, MobileNet and SegNet models and the CIFAR-10/100, ImageNet, and CamVid datasets.
The results showed that the proposed method can prevent accuracy degradation while achieving more effective compression than conventional methods by preferentially pruning the redundancy of these channels.
The results of the ablation study suggest that our approach effectively extracts information for pruning from attention statistics.
Analysis of attention statistics showed that there exist channels with a weak response to all classes and channels with a weak response to particular classes.
We believe that the attention mechanism can also be a useful technique for channel pruning.
\vspace{-15pt}
\section*{Acknowledgement}
\vspace{-8pt}
This paper is partly based on results obtained from a project commissioned by the New Energy and Industrial Technology Development Organization (NEDO).

\bibliography{references.bib}
\end{document}